\def\year{2019}\relax
\documentclass[letterpaper]{article} %DO NOT CHANGE THIS
\usepackage{aaai19}  %Required
\usepackage{times}  %Required
\usepackage{helvet}  %Required
\usepackage{courier}  %Required
\usepackage{url}  %Required
\usepackage{graphicx}  %Required
\graphicspath {img/}
\usepackage{amssymb,amsmath,mathtools}
\usepackage[ruled]{algorithm2e}
\usepackage{subfig}
\usepackage[LimeGreen]{xcolor}
\newcommand{\citet}[1]{\citeauthor{#1} ̃\shortcite{#1}}
\newcommand{\citep}{\cite}
\newcommand{\citealt}[1]{\citeauthor{#1} ̃\citeyear{#1}}

\begin{document}

\title{QUOTA: The Quantile Option Architecture for Reinforcement Learning}
\author{Shangtong Zhang$^1$~\thanks{Work done during an internship at Huawei}, Borislav Mavrin$^2$, Linglong Kong$^3$~\thanks{Work done during a sabbatical at Huawei}, Bo Liu$^4$, Hengshuai Yao$^2$\\
$^1$ Department of Computing Science, University of Alberta\\
$^2$ Reinforcement Learning for Autonomous Driving Lab, Noah's Ark Lab, Huawei \\
$^3$ Department of Mathematical and Statistical Sciences, University of Alberta\\ 
$^4$ Department of Computer Science and Software Engineering, Auburn University\\
\{shangtong.zhang, lkong\}@ualberta.ca, \{borislav.mavrin, hengshuai.yao\}@huawei.com, boliu@auburn.edu\\
}
\maketitle
\begin{abstract}
In this paper, we propose the Quantile Option Architecture (QUOTA) for exploration based on recent advances in distributional reinforcement learning (RL). In QUOTA, decision making is based on quantiles of a value distribution, not only the mean. QUOTA provides a new dimension for exploration via making use of both \textit{optimism} and \textit{pessimism} of a value distribution. We demonstrate the performance advantage of QUOTA in both challenging video games and physical robot simulators. 
\end{abstract}

\section{Introduction}

\noindent Mean of the return has been the center for reinforcement learning (RL) for a long time, and there have been many methods to learn a mean quantity (\citealt{sutton1988learning}; \citealt{watkins1992q}; \citealt{mnih2015human}). Thanks to the advances in distributional RL (\citealt{jaquette1973markov}; \citealt{bellemare2017distributional}), we are able to learn the full distribution, not only the mean, for a state-action value. Particularly, \citet{dabney2017distributional} used a set of quantiles to approximate this value distribution. However, the decision making in prevailing distributional RL methods is still based on the mean (\citealt{bellemare2017distributional}; \citealt{dabney2017distributional}; \citealt{barth2018distributed}; \citealt{qu2018nonlinear}). The main motivation of this paper is to answer the questions of how to make decision based on the full distribution and whether an agent can benefit for better exploration. In this paper, we propose the Quantile Option Architecture (QUOTA) for control. In QUOTA, decision making is based on all quantiles, not only the mean, of a state-action value distribution.

In traditional RL and recent distributional RL, an agent selects an action greedily with respect to the mean of the action values. In QUOTA, we propose to select an action greedily w.r.t. certain quantile of the action value distribution. A high quantile represents an \textit{optimistic} estimation of the action value, and action selection based on a high quantile indicates an optimistic exploration strategy. A low quantile represents a \textit{pessimistic} estimation of the action value, and action selection based on a low quantile indicates a pessimistic exploration strategy. (The two exploration strategies are related to risk-sensitive RL, which will be discussed later.) We first compared different exploration strategies in two Markov chains, where naive mean-based RL algorithms fail to explore efficiently as they cannot exploit the distribution information during training, which is crucial for efficient exploration. In the first chain, faster exploration is from a high quantile (i.e., an optimistic exploration strategy). However, in the second chain, exploration benefits from a low quantile (i.e., a pessimistic exploration strategy). Different tasks need different exploration strategies. Even within one task, an agent may still need different exploration strategies at different stages of learning. To address this issue, we use the option framework (\citealt{sutton1999between}). We learn a high-level policy to decide which quantile to use for action selection. In this way, different quantiles function as different options, and we name this special option the \textit{quantile option}. QUOTA adaptively selects a pessimistic and optimistic exploration strategy, resulting in improved exploration consistently across different tasks. 

We make two main contributions in this paper:
\begin{itemize}
\item First, we propose QUOTA for control in discrete-action problems, combining distributional RL with options. Action selection in QUOTA is based on certain quantiles instead of the mean of the state-action value distribution, and QUOTA learns a high-level policy to decide which quantile to use for decision making.
\item Second, we extend QUOTA to continuous-action problems. In a continuous-action space, applying quantile-based action selection is not straightforward. To address this issue, we introduce \textit{quantile actors}. Each quantile actor is responsible for proposing an action that maximizes one specific quantile of a state-action value distribution.
\end{itemize}

We show empirically QUOTA improves the exploration of RL agents, resulting in a performance boost in both challenging video games (Atari games) and physical robot simulators (Roboschool tasks)  

In the rest of this paper, we first present some preliminaries of RL. We then show two Markov chains where naive mean-based RL algorithms fail to explore efficiently. Then we present QUOTA for both discrete- and continuous-action problems, followed by empirical results. Finally, we give an overview of related work and closing remarks.

\section{Preliminaries}
We consider a Markov Decision Process (MDP) of a state space $\mathcal{S}$, an action space $\mathcal{A}$, a reward ``function'' $R: \mathcal{S} \times \mathcal{A} \rightarrow \mathbb{R}$, which we treat as a random variable in this paper, a transition kernel $p: \mathcal{S} \times \mathcal{A} \times \mathcal{S} \rightarrow [0, 1]$, and a discount ratio $\gamma \in [0, 1]$. We use $\pi: \mathcal{S} \times \mathcal{A} \rightarrow [0, 1]$ to denote a stochastic policy. We use $Z^\pi(s, a)$ to denote the random variable of the sum of the discounted rewards in the future, following the policy $\pi$ and starting from the state $s$ and the action $a$. We have $Z^\pi(s, a) \doteq \sum_{t=0}^\infty \gamma^t R(S_t, A_t)$, where $S_0 = s, A_0 = a$ and $S_{t+1} \sim p(\cdot | S_t, A_t), A_t \sim \pi(\cdot| S_t)$. The expectation of the random variable $Z^\pi(s, a)$ is $Q^\pi(s, a) \doteq \mathbb{E}_{\pi, p, R}[Z^\pi(s, a)]$, which is usually called the state-action value function. We have the Bellman equation $$Q^\pi(s, a) = \mathbb{E}[R(s, a)] + \gamma \mathbb{E}_{s^\prime \sim p(\cdot| s, a), a^\prime \sim \pi(\cdot | s^\prime)}[Q^\pi(s^\prime, a^\prime)]$$

In a RL problem, we are usually interested in finding an optimal policy $\pi^*$ such that $Q^{\pi^*}(s, a) \geq Q^\pi(s, a) \, \forall \, (\pi, s, a)$. All the possible optimal policies share the same (optimal) state action value function $Q^*$. This $Q^*$ is the unique fixed point of the Bellman optimality operator $\mathcal{T}$(\citealt{bellman2013dynamic})
\begin{align*}
\mathcal{T}Q(s, a) \doteq \mathbb{E}[R(s, a)] + \gamma \mathbb{E}_{s^\prime \sim p}[\max_{a^\prime} Q(s^\prime, a^\prime)]
\end{align*}
With tabular representation, we can use Q-learning (\citealt{watkins1992q}) to estimate $Q^*$. The incremental update per step is 
\begin{align}
\label{eq:q}
\Delta Q(s_t, a_t) \propto r_{t+1} + \gamma \max_{a}Q(s_{t+1}, a) - Q(s_t, a_t)
\end{align}
where the quadruple $(s_t, a_t, r_{t+1}, s_{t+1})$ is a transition. There are lots of research and algorithms extending Q-learning to linear function approximation (\citealt{sutton2018reinforcement}; \citealt{szepesvari2010algorithms}). In this paper, we focus on Q-learning with neural networks. \citet{mnih2015human} proposed Deep-Q-Network (DQN), where a deep convolutional neural network $\theta$ is used to parameterize $Q$. At every time step, DQN performs stochastic gradient descent to minimize
\begin{align*}
\frac{1}{2}(r_{t+1} + \gamma \max_a Q_{\theta^-}(s_{t+1}, a) - Q_\theta(s_t, a_t)) ^ 2
\end{align*}
where the quadruple $(s_t, a_t, r_{t+1}, s_{t+1})$ is a transition sampled from the replay buffer (\citealt{lin1992self}) and $\theta^-$ is the target network (\citealt{mnih2015human}), which is a copy of $\theta$ and is synchronized with $\theta$ periodically. To speed up training and reduce the required memory of DQN, \citet{mnih2016asynchronous} further proposed the $n$-step asynchronous Q-learning with multiple workers (detailed in Supplementary Material), where the loss function at time step $t$ is
\begin{align*}
\frac{1}{2}(\sum_{i=1}^n \gamma^{i-1}r_{t+i} + \gamma^n \max_a Q_{\theta^-}(s_{t+n}, a) - Q_\theta(s_t, a_t)) ^ 2
\end{align*}

\subsection{Distributional RL}
Analogous to the Bellman equation of $Q^\pi$, \citet{bellemare2017distributional} proposed the distributional Bellman equation for the state-action value distribution $Z^\pi$ given a policy $\pi$ in the policy evaluation setting,
\begin{align*}
Z^\pi(s, a) \stackrel{D}{=} R(s, a) + \gamma Z^\pi(s^\prime, a^\prime) \\
s^\prime \sim p(\cdot|s, a) , a^\prime \sim \pi(\cdot|s)
\end{align*}
where $X \stackrel{D}{=} Y$ means the two random variables $X$ and $Y$ are distributed according to the same law. \citet{bellemare2017distributional} also proposed a distributional Bellman optimality operator for control,
\begin{align*}
\mathcal{T}Z(s, a) \doteq R(s, a) + \gamma Z(s^\prime, \arg\max_{a^\prime}\mathbb{E}_{p, R}[Z(s^\prime, a^\prime)]) \\
s^\prime \sim p(\cdot|s, a)
\end{align*}
When making decision, the action selection is still based on the expected state-action value (i.e., $Q$). Since we have the optimality, now we need an representation for $Z$. \citet{dabney2017distributional} proposed to approximate $Z(s, a)$ by a set of quantiles. The distribution of $Z$ is represented by a uniform mix of $N$ supporting quantiles:
\begin{align*}
Z_\theta(s, a) \doteq \frac{1}{N}\sum_{i=1}^N \delta_{q_i(s, a; \theta)}
\end{align*}
where $\delta_x$ denote a Dirac at $x \in \mathbb{R}$, and each $q_i$ is an estimation of the quantile corresponding to the quantile level (a.k.a. quantile index) $\hat{\tau}_i \doteq \frac{\tau_{i-1} + \tau_i}{2}$ with $\tau_i \doteq \frac{i}{N}$ for $0 \leq i \leq N$. The state-action value $Q(s, a)$ is then approximated by $\frac{1}{N}\sum_{i=1}^N q_i(s, a)$. Such approximation of a distribution is referred to as quantile approximation. Those quantile estimations (i.e., $\{q_i\}$) are trained via the Huber quantile regression loss (\citealt{huber1964robust}). To be more specific, at time step $t$ the loss is 
\begin{align*}
\frac{1}{N}\sum_{i=1}^N \sum_{i^\prime=1}^N\Big[\rho_{\hat{\tau}_i}^\kappa\big(y_{t, i^\prime} - q_i(s_t, a_t)\big)\Big]
\end{align*}
where $y_{t, i^\prime} \doteq r_t + \gamma q_{i^\prime}\big(s_{t+1}, \arg\max_{a^\prime}\sum_{i=1}^N q_i(s_{t+1}, a^\prime)\big)$ and $\rho_{\hat{\tau}_i}^\kappa(x) \doteq |\hat{\tau}_i - \mathbb{I}\{x < 0\}|\mathcal{L}_\kappa(x)$, where $\mathbb{I}$ is the indicator function and $\mathcal{L}_\kappa$ is the Huber loss,
\begin{align*}
\mathcal{L}_\kappa(x) \doteq \begin{cases}
\frac{1}{2}x^2 & \text{if } x \leq \kappa \\
\kappa(|x| - \frac{1}{2}\kappa) & \text{otherwise}
\end{cases}
\end{align*}
The resulting algorithm is the Quantile Regression DQN (QR-DQN). QR-DQN also uses experience replay and target network similar to DQN. \citet{dabney2017distributional} showed that quantile approximation has better empirical performance than previous categorical approximation (\citealt{bellemare2017distributional}). More recently, \citet{dabney2018implicit} approximated the distribution by learning a quantile function directly with the Implicit Quantile Network, resulting in further performance boost. Distributional RL has enjoyed great success in various domains (\citealt{bellemare2017distributional}; \citealt{dabney2017distributional}; \citealt{hessel2017rainbow}; \citealt{barth2018distributed}; \citealt{dabney2018implicit}).

\subsection{Deterministic Policy}
\citet{silver2014deterministic} used a deterministic policy $\mu: \mathcal{S} \rightarrow \mathcal{A}$ for continuous control problems with linear function approximation, and \citet{lillicrap2015continuous} extended it with deep networks, resulting in the Deep Deterministic Policy Gradient (DDPG) algorithm. DDPG is an off-policy algorithm. It has an actor $\mu$ and a critic $Q$, parameterized by $\theta^\mu$ and $\theta^Q$ respectively. At each time step, $\theta^Q$ is updated to minimize 
\begin{align*}
\frac{1}{2}(r_{t + 1} + \gamma Q(s_{t+1}, \mu(s_{t+1})) - Q(s_t, a_t))^2
\end{align*}
And the policy gradient for $\theta^\mu$ in DDPG is 
\begin{align*}
\nabla_a Q(s_t, a)|_{a=\mu(s_t)} \nabla_{\theta^\mu}\mu(s_t)
\end{align*}
This gradient update is from the chain rule of gradient ascent w.r.t. $Q(s_t, \mu(s_t))$, where $\mu(s_t)$ is interpreted as an approximation to $\arg\max_a Q(s_t, a)$. \citet{silver2014deterministic} provided policy improvement guarantees for this gradient.

\subsection{Option}
An option (\citealt{sutton1999between}) is a temporal abstraction of action. Each option $\omega \in \Omega$ is a triple $(\mathcal{I}_\omega, \pi_\omega, \beta_\omega)$, where $\Omega$ is the option set. We use $\mathcal{I}_\omega \subseteq \mathcal{S}$ to denote the initiation set for the option $\omega$, describing where the option $\omega$ can be initiated. We use $\pi_\omega: \mathcal{S} \times \mathcal{A} \rightarrow [0, 1]$ to denote the intra-option policy for $\omega$. Once the agent has committed to the option $\omega$, it chooses an action based on $\pi_\omega$. We use $\beta_\omega: \mathcal{S} \rightarrow [0, 1]$ to denote the termination function for the option $\omega$. At each time step $t$, the option $\omega_{t-1}$ terminates with probability $\beta_{\omega_{t-1}}(S_t)$. In this paper, we consider the call-and-return option execution model (\citealt{sutton1999between}), where an agent commits to an option $\omega$ until $\omega$ terminates according to $\beta_\omega$.

The option value function $Q_\Omega(s, \omega)$ is used to describe the utility of an option $\omega$ at state $s$, and we can learn this function via Intra-option Q-learning (\citealt{sutton1999between}). The update is
\begin{align*}
&Q_\Omega(s_t, \omega_t) \leftarrow Q_\Omega(s_t, \omega_t) + \\
& \quad \alpha \Big(r_{t+1} + \gamma \big(\beta_{\omega_t}(s_{t+1})\max_{\omega^\prime}Q_\Omega(s_{t+1}, \omega^\prime) \\
& \quad + (1 - \beta_{\omega_t}(s_{t+1}))Q_\Omega(s_{t+1}, \omega_t)\big) - Q_\Omega(s_t, \omega_t)\Big)
\end{align*}
where $\alpha$ is a step size and $(s_t, a_t, r_{t+1}, s_{t+1})$ is a transition in the cycle that the agent is committed to the option $\omega_t$.

\section{A Failure of Mean}

\begin{figure*}[h]
\centering
\subfloat[]{\includegraphics[width=0.33\textwidth]{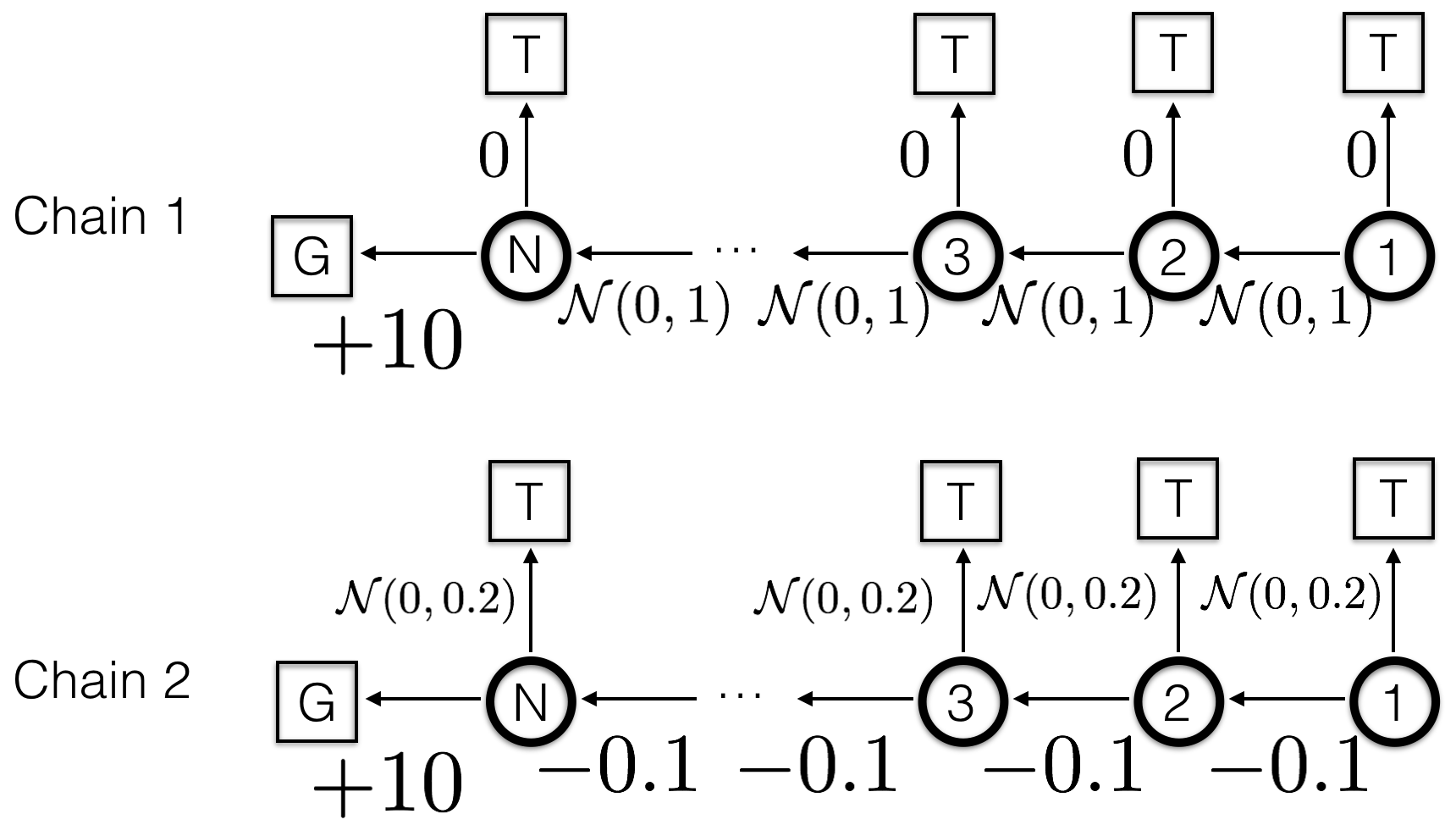}}
\subfloat[Chain 1]{\includegraphics[width=0.33\textwidth]{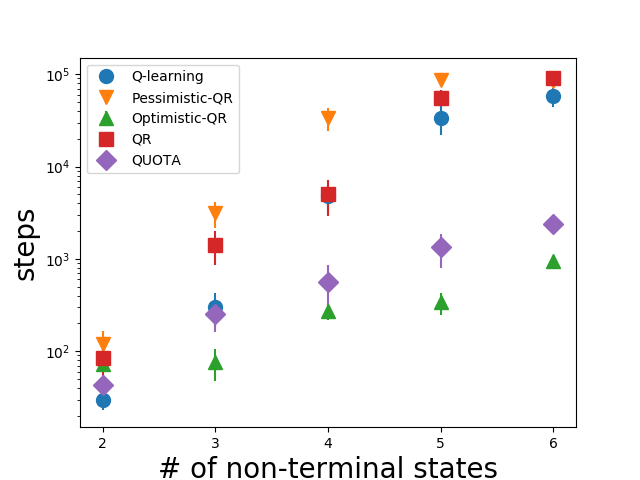}}
\subfloat[Chain 2]{\includegraphics[width=0.33\textwidth]{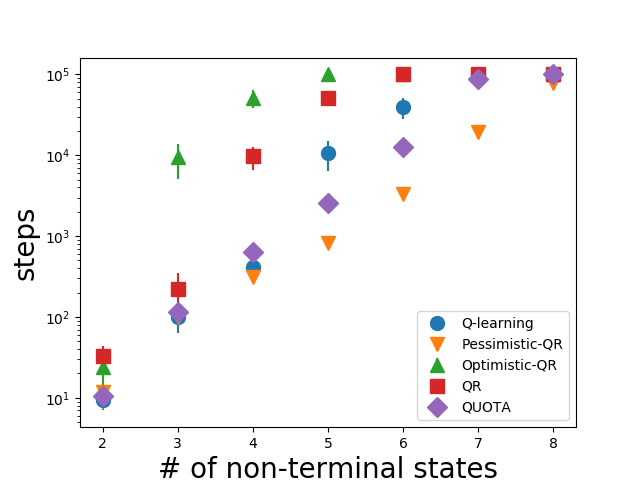}}
\caption{\label{fig:chains} (a) two Markov chains illustrating the inefficiency of mean-based decision making. (b)(c) show the required steps to find the optimal policy for each algorithm v.s. the chain length for Chain~1 and Chain~2 respectively. The required steps are averaged over 10 trials, and standard errors are plotted as error bars. For each trial, the maximum step is $10^5$.}
\end{figure*}

We now present two simple Markov chains (Figure~\ref{fig:chains}) to illustrate mean-based RL algorithms can fail to explore efficiently. 

Chain 1 has $N$ non-terminal states and two actions \{\textit{LEFT}, \textit{UP}\}. The agent starts at the state 1 in each episode. The action \textit{UP} will lead to episode ending immediately with reward 0. The action \textit{LEFT} will lead to the next state with a reward sampled from a normal distribution $\mathcal{N}(0, 1)$. Once the agent runs into the \textit{G} terminal state, the episode will end with a reward $+10$. There is no discounting. The optimal policy is always moving left. 

We first consider tabular Q-learning with $\epsilon$-greedy exploration. To learn the optimal policy, the agent has to reach the \textit{G} state first. Unfortunately, this is particularly difficult for Q-learning. The difficulty comes from two aspects. First, due to the $\epsilon$-greedy mechanism, the agent sometimes selects \textit{UP} by randomness. Then the episode will end immediately, and the agent has to wait for the next episode. Second, before the agent reaches the \textit{G} state, the expected return of either \textit{LEFT} or \textit{UP} at any state is 0. So the agent cannot distinguish between the two actions under the mean criterion because the expected returns are the same. As a result, the agent cannot benefit from the $Q$ value estimation, a mean, at all. 

Suppose now the agent learns the distribution of the returns of \textit{LEFT} and \textit{UP}. Before reaching the state \textit{G}, the learned action-value distribution of \textit{LEFT} is a normal distribution with a mean 0. A high quantile level of this distribution is greater than 0, which is an optimistic estimation. If the agent behaves according to this optimistic estimation, it can quickly reach the state \textit{G} and find the optimal policy.

Chain 2 has the same state space and action space as Chain 1. However, the reward for \textit{LEFT} is now $-0.1$ except that reaching the \textit{G} state gives a reward $+10$. The reward for \textit{UP} is sampled from $\mathcal{N}(0, 0.2)$. There is no discounting. When $N$ is small, the optimal policy is still always moving left. Before reaching $G$, the estimation of the expected return of \textit{LEFT} for any non-terminal state is less than 0, which means a Q-learning agent would prefer \textit{UP}. This preference is bad for this chain as it will lead to episode ending immediately, which prevents further exploration.

We now present some experimental results of four algorithms in the two chains: Q-learning, quantile regression Q-learning (QR, the tabular version of QR-DQN), optimistic quantile regression Q-learning (O-QR), and pessimistic quantile regression Q-learning (P-QR). The O-QR / P-QR is the same as QR except that the behavior policy is always derived from a high / low quantile, not the mean, of the state-action value distribution. We used $\epsilon$-greedy behavior policies for all the above algorithms. We measured the steps that each algorithm needs to find the optimal policy. The agent is said to have found the optimal policy at time $t$ if and only if the policy derived from the $Q$ estimation (for Q-learning) or the mean of the $Z$ estimation (for the other algorithms) at time step $t$ is to move left at all non-terminal states.

All the algorithms were implemented with a tabular state representation. $\epsilon$ was fixed to $0.1$.  For quantile regression, we used 3 quantiles. We varied the chain length and tracked the steps that an algorithm needed to find the optimal policy. Figure~\ref{fig:chains}b and Figure~\ref{fig:chains}c show the results in Chain 1 and Chain 2 respectively. Figure~\ref{fig:chains}b shows the best algorithm for Chain~1 was O-QR, where a high quantile is used to derive the behavior policy, indicating an optimistic exploration strategy. P-QR performed poorly with the increase of the chain length. Figure~\ref{fig:chains}c shows the best algorithm for Chain~2 was P-QR, where a low quantile is used to derive the behavior policy, indicating a pessimistic exploration strategy. O-QR performed poorly with the increase of the chain length. The mean-based algorithms (Q-learning and QR) performed poorly in both chains. Although QR did learn the distribution information, it did not use that information properly for exploration. 

The results in the two chains show that quantiles influence exploration efficiency, and quantile-based action selection can improve exploration if the quantile is properly chosen. The results also demonstrate that different tasks need different quantiles. No quantile is the best universally. As a result, a high-level policy for quantile selection is necessary. 

\section{The Quantile Option Architecture}

We now introduce QUOTA for discrete-action problems. We have $N$ quantile estimations $\{q_i\}_{i=1, \dots, N}$ for quantile levels $\{\hat{\tau}_i\}_{i=1, \dots, N}$. We construct $M$ options $(M < N)$. For simplicity, in this paper all the options share the same initiation set $\mathcal{S}$ and the same termination function $\beta$, which is a constant function. We use $\pi^j$ to indicate the intra-option policy of an option $\omega^j, j = 1, \dots, M$. $\pi^j$ proposes actions based on the mean of the $j$-th window of $K$ quantiles, where $K \doteq N/M$. (We assume $N$ is divisible by $M$ for simplicity.) Here $K$ represents a window size and we have $M$ windows in total. To be more specific, in order to compose the $j$-th option, we first define a state-action value function $Q_{j|K}$ by averaging a local window of $K$ quantiles:
\begin{align*}
Q_{j|K}(s, a) \doteq \frac{1}{K}\sum_{k= (j-1) K + 1}^{(j-1)K + K}q_k(s, a)
\end{align*}
We then define the intra-option policy $\pi^j$ of the $j$-th option $\omega^j$ to be an $\epsilon$-greedy policy with respect to $Q_{j|K}$. Here we compose an option with a window of quantiles, instead of a single quantile, to increase stability. It appears to be a mean form, but it is not the mean of the full state-action value distribution. QUOTA learns the option-value function $Q_\Omega$ via Intra-option Q-learning for option selection. The quantile estimations $\{q_i\}$ is learned via QR-DQN. 

To summarize, at each time step $t$, we reselect a new option with probability $\beta$ and continue executing the previous option with probability $1 - \beta$. The reselection of the new option is done via a $\epsilon_\Omega$-greedy policy derived from $Q_\Omega$, where $\epsilon_\Omega$ is the random option selection probability. Once the current option $\omega_t$ is determined, we then select an action $a_t$ according to the intra-option policy of $\omega_t$. The pseudo code of QUOTA is provided in Supplementary Material. 

\subsection{QUOTA for Continuous Control}
\subsubsection{QR-DDPG} Quantile Regression DDPG (QR-DDPG, detailed in Supplementary Material) is a new algorithm by modifying DDPG's critic. Instead of learning $Q$, we learn the distribution $Z$ directly in the same manner as the discrete-action controller QR-DQN. \citet{barth2018distributed} also learned $Z$ instead of $Q$. However, they parameterized $Z$ through categorical approximation and mixed Gaussian approximation. To our best knowledge, QR-DDPG is the first to parameterize $Z$ with quantile estimations in continuous-action problems. We use QR-DDPG and DDPG as baselines.

\subsubsection{QUOTA} Given the distribution $Z$ of the state-action value approximated by quantiles, the main question now is how to select an action according to certain quantile. For a finite discrete action space, action selection is done according to an $\epsilon$-greedy policy with respect to $Q_{j|K}$, where we need to iterate through all possible actions in the whole action space. To get the action that maximizes a quantile of a distribution in continuous-action problems, we perform gradient ascent for different intra-option policies in analogy to DDPG. We have $M$ options $\{\omega^j\}_{j=1,\dots,M}$. The intra-option policy for the option $\omega^j$ is a deterministic mapping $\mu_j: \mathcal{S} \rightarrow \mathcal{A}$. We train $\mu_j$ to approximate the greedy action $\arg\max_a Q_{j|K}(s, a)$ via gradient ascent. To be more specific, the gradient for $\mu^j$ (parameterized by $\phi$) at time step $t$ is
\begin{align*}
\nabla_a Q_{j|K}(s_t, a)|_{a = \mu^j(s_t)} \nabla_\phi \mu^j(s_t)
\end{align*}
To compute the update target for the critic $Z$, we also need one more actor $\mu^0$ to maximize the mean of the distribution (i.e., $\frac{1}{N}\sum_{i=1}^N q_i$) as it is impossible to iterate through all the actions in a continuous-action space. Note $\mu^0$ is the same as the actor of QR-DDPG. We augment QUOTA's option set with $\mu^0$, giving $M + 1$ options. We name $\mu^j$ the $j$-\textit{th quantile actor} ($j=1, \dots M)$. QUOTA for continuous-action problems is detailed in Supplementary Material.

\section{Experiments}

We designed experiments to study whether QUOTA improves exploration and can scale to challenging tasks. All the implementations are made publicly available.~\footnote{https://github.com/ShangtongZhang/DeepRL}

\subsection{Does QUOTA improve exploration?}
We benchmarked QUOTA for discrete-action problems in the two chains. We used a tabular state representation and three quantiles to approximate $Z$ as previous experiments. Both $\epsilon$ and $\epsilon_\Omega$ were fixed at 0.1, $\beta$ was fixed at 0, which means an option never terminated and lasted for a whole episode. The results are reported in Figures~\ref{fig:chains}b and \ref{fig:chains}c. QUOTA consistently performed well in both chains. 

Although QUOTA did not achieve the best performance in the two chains, it consistently reached comparable performance level with the best algorithm in both chains. Not only the best algorithm in each chain was designed with certain domain knowledge, but also the best algorithm in one chain performed poorly in the other chain. We do not expect QUOTA to achieve the best performance in both chains because it has not used chain-specific knowledge. Those mean-based algorithms (Q-learning and QR) consistently performed poorly in both chains. QUOTA achieved more efficient exploration than Q-learning and QR. 

\subsection{Can QUOTA scale to challenging tasks?}
To verify the scalability of QUOTA, we evaluated QUOTA in both Arcade Learning Environment (ALE) (\citealt{bellemare2013arcade}) and Roboschool~\footnote{https://blog.openai.com/roboschool/}, both of which are general-purpose RL benchmarks. 

\subsubsection{Arcade Learning Environment}
We evaluated QUOTA in the 49 Atari games from ALE as \citet{mnih2015human}. Our baseline algorithm is QR-DQN. We implemented QR-DQN with multiple workers (\citealt{mnih2016asynchronous}; \citealt{clemente2017efficient}) and an $n$-step return extension (\citealt{mnih2016asynchronous}; \citealt{hessel2017rainbow}), resulting in reduced wall time and memory consumption compared with an experience-replay-based implementation. We also implemented QUOTA in the same way. Details and more justification for this implementation can be found in Supplementary Material.

% The training of Atari games with experience replay is time-consuming and memory-consuming. Asynchronous methods (\citealt{mnih2016asynchronous}) replace the experience replay mechanism by asynchronous workers, resulting in reduced wall time and memory. More recently, synchronous workers are used in place of asynchronous workers to utilize a modern GPU (\citealt{clemente2017efficient}). Our baseline algorithm is QR-DQN. However, we implemented a QR-DQN with synchronous workers and an $n$-step extension like the $n$-step Q-learning, resulting in reduced training time. 

We used the same network architecture as \citet{dabney2017distributional} to process the input pixels. For QUOTA, we added an extra head to produce the option value $Q_\Omega$ after the second last layer of the network. For both QUOTA and QR-DQN, we used 16 synchronous workers, and the rollout length is 5, resulting in a batch size 80. We trained each agent for 40M steps with frameskip 4, resulting in 160M frames in total. We used an RMSProp optimizer with an initial learning rate $10^{-4}$. The discount factor is 0.99. The $\epsilon$ for action selection was linearly decayed from 1.0 to 0.05 in the first 4M training steps and remained 0.05 afterwards. All the hyper-parameter values above were based on an $n$-step Q-learning baseline with synchronous workers from \citet{farquhar2018treeqn}, and all our implementations inherited these hyper-parameter values. We used 200 quantiles to approximate the distribution and set the Huber loss parameter $\kappa$ to 1 as used by \citet{dabney2017distributional}. We used 10 options in QUOTA ($M=10$) , and $\epsilon_\Omega$ was linearly decayed from 1.0 to 0 during the 40M training steps. $\beta$ was fixed at 0.01. We tuned $\beta$ from $\{0, 0.01, 0.1, 1\}$ in the game Freeway. The schedule of $\epsilon_\Omega$ was also tuned in Freeway. 

\begin{figure*}
\centering
\subfloat[Final performance improvement of QUOTA over QR-DQN]{\includegraphics[width=1.0\textwidth]{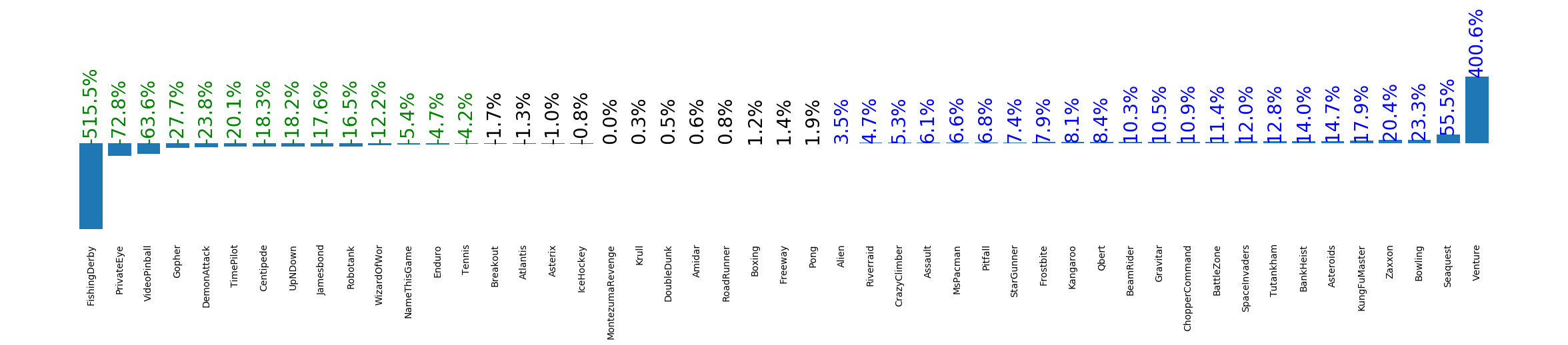}} \\
\subfloat[Cumulative rewards improvement of QUOTA over QR-DQN]{\includegraphics[width=1.0\textwidth]{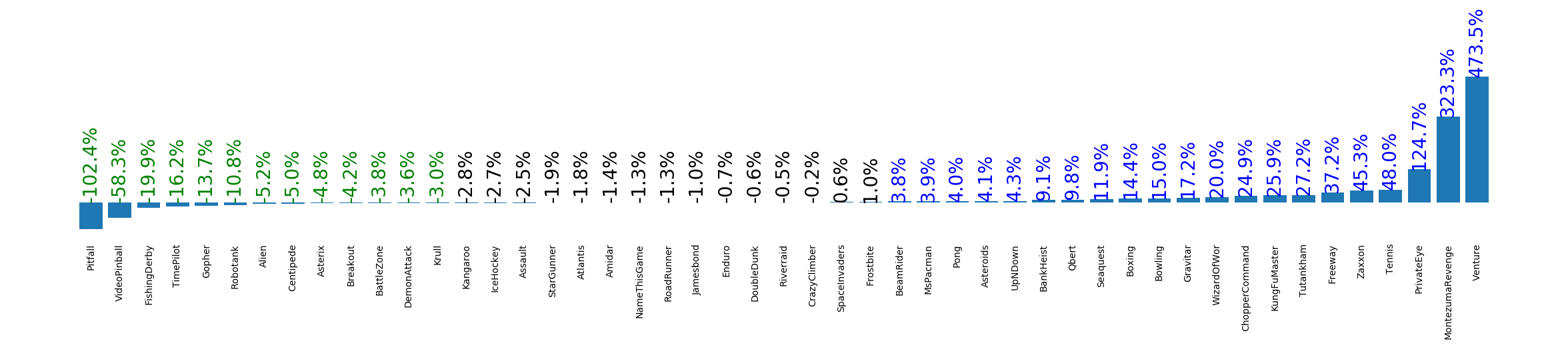}}
\caption{\label{fig:atari} Improvements in Atari games. Numbers indicate improvement and are computed as $(\text{score}_\text{QUOTA} - \text{score}_\text{QR-DQN}) / |\text{score}_\text{QR-DQN}|$, where scores in (a) are the final performance (averaged episode return of the last 1,000 episodes at the end of training), and scores in (b) are cumulative rewards during training. All the scores are averaged across 3 independent runs. Bars above / below horizon indicate performance gain / loss.} 
\end{figure*}

We measured the final performance at the end of training (i.e., the mean episode return of the last 1,000 episodes) and the cumulative rewards during training. The results are reported in Figure~\ref{fig:atari}. In terms of the final performance / cumulative rewards, QUOTA outperformed QR-DQN in 23 / 21 games and underperformed QR-DQN in 14 / 13 games. Here we only considered performance change larger than 3\%. Particularly, in the 10 most challenging games (according to the scores of DQN reported in \citet{mnih2015human}), QUOTA achieved a 97.2\% cumulative reward improvement in average.

In QUOTA, randomness comes from both $\epsilon$ and $\epsilon_\Omega$. However, in QR-DQN, randomness only comes from $\epsilon$. So QUOTA does have more randomness (i.e., exploration) than QR-DQN when $\epsilon$ is the same. To make it fair, we also implemented an alternative version of QR-DQN, referred to as QR-DQN-Alt, where all the hyper-parameter values were the same except that $\epsilon$ was linearly decayed from 1.0 to 0 during the whole 40M training steps like $\epsilon_\Omega$. In this way, QR-DQN-Alt had a comparable amount of exploration with QUOTA. 

We also benchmarked QR-DQN-Alt in the 49 Atari games. In terms of the final performance / cumulative rewards, QUOTA outperformed QR-DQN-Alt in 27 / 42 games and underperformed QR-DQN-Alt in 14 / 5 games, indicating a naive increase of exploration via tuning the $\epsilon$ dose not guarantee a performance boost. 

All the original learning curves and scores are reported in Supplementary Material. 

\subsubsection{Roboschool}
Roboschool is a free port of Mujoco~\footnote{http://www.mujoco.org/} by OpenAI, where a state contains joint information of a robot and an action is a multi-dimensional continuous vector. We consider DDPG and QR-DDPG as our baselines, implemented with experience replay. For DDPG, we used the same hyper-parameter values and exploration noise as \citet{lillicrap2015continuous}, except that we found replacing $ReLU$ and $L_2$ regularizer with $tanh$ brought in a performance boost in our Roboschool tasks. Our implementations of QUOTA and QR-DDPG inherited these hyper-parameter values. For QR-DDPG, we used 20 output units after the second last layer of the critic network to produce 20 quantile estimations for the action value distribution. For QUOTA, we used another two-hidden-layer network with 400 and 300 hidden units to compute $Q_\Omega$. (This network is the same as the actor network in DDPG and QR-DDPG.) We used 6 options in total, including one option that corresponds to the actor maximizing the mean value. $\epsilon_\Omega$ was linearly decayed from 1.0 to 0 in the whole 1M training steps. $\beta$ was fixed at 1.0, which means we reselected an option at every time step. We tuned $\beta$ from $\{0, 0.01, 0.1, 1\}$ in the game Ant. We trained each algorithm for 1M steps and performed 20 deterministic evaluation episodes every 10K training steps. 

\begin{figure*}
\centering
\includegraphics[width=1.0\textwidth]{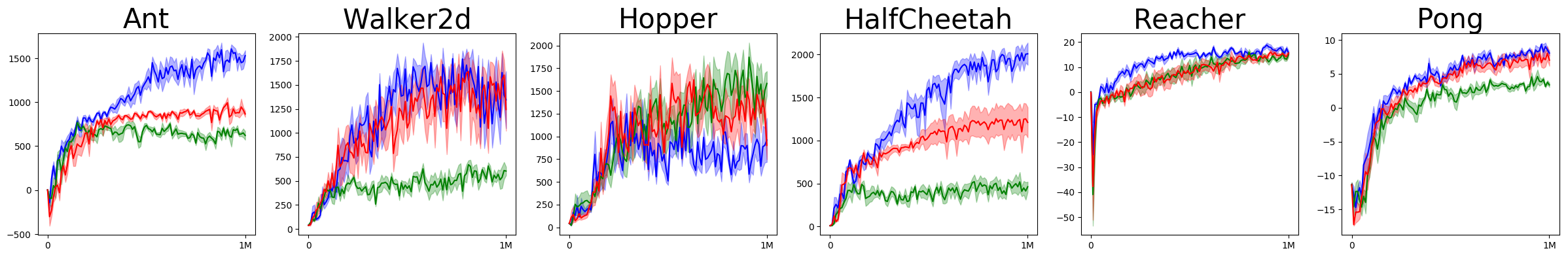}
\caption{\label{fig:roboschool}Evaluation curves of Roboschool tasks. The x-axis is training steps, and the y-axis is score. Blue lines indicate {\color{blue} QUOTA}, green lines indicate {\color{green} DDPG}, and red lines indicate {\color{red} QR-DDPG}. The curves are averaged over 5 independent runs, and standard errors are plotted as shadow.}
\end{figure*}

The results are reported in Figure~\ref{fig:roboschool}. QUOTA demonstrated improved performance over both DDPG and QR-DDPG in 5 out of the 6 tasks. For the other six tasks in Roboschool, all the compared algorithms had large variance and are hard to compare. Those results are reported in Supplementary Material.

\subsubsection{Visualization}
\begin{figure*}
\centering
\includegraphics[width=1.0\textwidth]{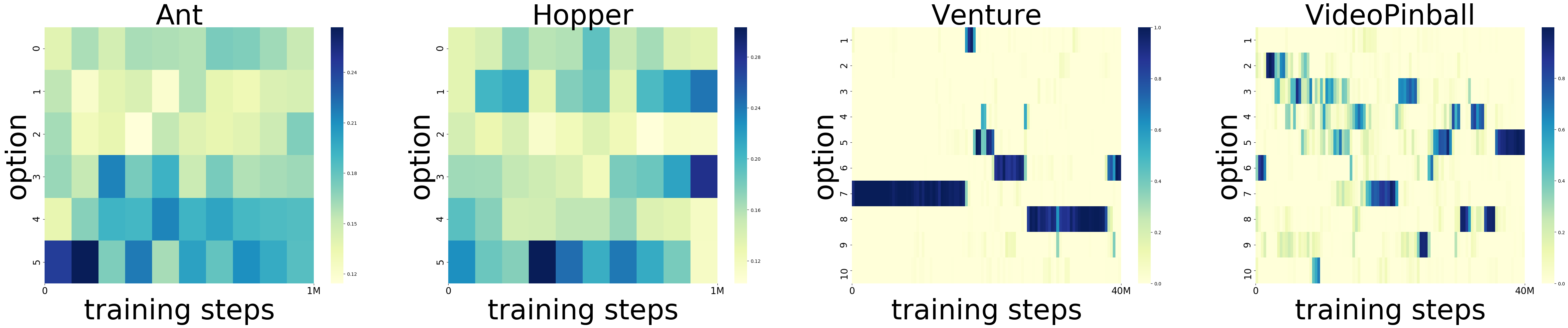}
\caption{\label{fig:heatmap} Frequency of the greedy option w.r.t. $Q_\Omega$. The color represents the frequency that an option was proposed by $Q_\Omega$ in different training stages. The frequencies in each column sum to 1. The darker a grid is, the more frequent the option is proposed at that time.}
\end{figure*}

To understand option selection during training, we plot the frequency of the greedy options according to $Q_\Omega$ in different stages of training in Figure~\ref{fig:heatmap}. At different training stage, $Q_\Omega$ did propose different quantile options. The quantile option corresponding to mean or median did not dominate the training. In fact, during training, the mean-maximizing options was rarely proposed by the high-level policy. This indicates that the traditional mean-centered action selection (adopted in standard RL and prevailing distributional RL) can be improved by quantile-based exploration.

\section{Related Work}

There have been many methods for exploration in model-free setting based on the idea of \textit{optimism in the face of uncertainty} (\citealt{lai1985asymptotically}). Different approaches are used to achieve this optimism, e.g., count-based methods (\citealt{auer2002using}; \citealt{kocsis2006bandit}; \citealt{bellemare2016unifying}; \citealt{ostrovski2017count}; \citealt{tang2017exploration}) and Bayesian methods (\citealt{kaufmann2012bayesian}; \citealt{chen2017ucb}; \citealt{o2017uncertainty}). Those methods make use of \textit{optimism} of the \textit{parametric uncertainty} (i.e., the uncertainty from estimation). In contrast, QUOTA makes use of both \textit{optimism} and \textit{pessimism} of the \textit{intrinsic uncertainty} (i.e., the uncertainty from the MDP itself). Recently, \citet{moerland2017efficient} combined the two uncertainty via the Double Uncertain Value Network and demonstrated performance boost in simple domains.

There is another line of related work in risk-sensitive RL. Classic risk-sensitive RL is also based on the intrinsic uncertainty, where a utility function (\citealt{morgenstern1953theory}) is used to distort a value distribution, resulting in risk-averse or risk-seeking policies (\citealt{howard1972risk}; \citealt{marcus1997risk}; \citealt{chow2014algorithms}). The expectation of the utility-function-distorted valued distribution can be interpreted as a weighted sum of quantiles (\citealt{dhaene2012remarks}), meaning that quantile-based action selection implicitly adopts the idea of utility function. Particularly, \citet{morimura2012parametric} used a small quantile for control, resulting in a safe policy in the cliff grid world. \citet{chris2017particle} employed an exponential utility function over the parametric uncertainty, also resulting in a performance boost in the cliff grid world. Recently, \citet{dabney2018implicit} proposed various risk-sensitive policies by applying different utility functions to a learned quantile function. The high-level policy $Q_\Omega$ in QUOTA can be interpreted as a special utility function, and the optimistic and pessimistic exploration in QUOTA can be interpreted as risk-sensitive policies. However, the ``utility function'' $Q_\Omega$ in QUOTA is formalized in the option framework, which is \textit{learnable}. We do not need extra labor to pre-specify a utility function. 

Moreover, \citet{tang2018exploration} combined Bayesian parameter updates with distributional RL for efficient exploration. However, improvements were demonstrated only in simple domains.

\section{Closing Remarks}
QUOTA achieves an on-the-fly decision between optimistic and pessimistic exploration, resulting in improved performance in various domains. QUOTA provides a new dimension for exploration. In this optimism-pessimism dimension, an agent may be able to find an effective exploration strategy more quickly than in the original action space. QUOTA provides an option-level exploration. 

At first glance, QUOTA introduces three extra hyper-parameters, i.e., the number of options $M$, the random option probability $\epsilon_\Omega$, and the termination probability $\beta$. For $M$, we simply used 10 and 5 options for Atari games and Roboschool tasks respectively. For $\epsilon_\Omega$, we used a linear schedule decaying from 1.0 to 0 during the whole training steps, which is also a natural choice. We do expect performance improvement if $M$ and $\epsilon_\Omega$ are further tuned. For $\beta$, we tuned it in only two environments and used the same value for all the other environments. Involved labor effort was little. Furthermore, $\beta$ can also be learned directly in an end-to-end training via the termination gradient theorem (\citealt{bacon2017option}). We leave this for future work.

Bootstrapped-DQN (BDQN, \citealt{osband2016deep}) approximated the distribution of the expected return via a set of $Q$ heads. At the beginning of each episode, BDQN uniformly samples one head and commits to that head during the whole episode. This uniform sampling and episode-long commitment is crucial to the deep exploration of BDQN and inspires us to set $\epsilon_\Omega = 1$ and $\beta = 0$. However, this special configuration only improved performance in certain tasks. Each head in BDQN is an estimation of $Q$ value. All heads are expected to converge to the optimal state action value at the end of training. As a result, a simple uniform selection over the heads in BDQN does not hurt performance. However, in QUOTA, each head (i.e., quantile option) is an estimation of a quantile of $Z$. Not all quantiles are useful for control. A selection among quantile options is necessary. One future work is to combine QUOTA and BDQN or other parametric-uncertainty-based algorithms (e.g., count-based methods) by applying them to each quantile estimation. 

% {\small \bibliography{../ref.bib}}
% \bibliographystyle{aaai}

\newpage
\onecolumn
\section{Supplementary Material}

\subsection{QUOTA for Discrete-Action Problems}

\begin{algorithm}[ht]
\DontPrintSemicolon
\textbf{Input:}\\
$\epsilon$: random action selection probability \\ 
$\epsilon_\Omega$: random option selection probability \\
$\beta$: option termination probability \\
$\{q_i\}_{i=1,\dots,N}$: quantile estimation functions, parameterized by $\theta$ \\
$Q_\Omega$: an option value function, parameterized by $\psi$  \\ 
\textbf{Output:}\\
parameters $\theta$ and $\psi$ \\
\hrulefill\\
\For{each time step t} {
Observe the state $s_t$ \\
Select an option $\omega_t$, \\
$\omega_t \leftarrow \begin{cases}
\omega_{t-1} & \text{w.p. } 1 - \beta \\
\text{random option} & \text{w.p. } \beta \epsilon_\Omega \\
\arg\max_{\omega \in \{\omega^j\}} Q_\Omega(s_t, \omega)  & \text{w.p. } \beta (1 - \epsilon_\Omega) 
\end{cases}$ \\
Select an action $a_t$ (assuming $\omega_t$ is the $j$-th option), \\ 
$a_t \leftarrow \begin{cases}
\text{random action} & \text{w.p. } \epsilon \\
\arg\max_{a \in \mathcal{A}} \sum_{k= (j-1) K + 1}^{(j-1)K + K}q_k(s_t, a) & \text{w.p. } 1 - \epsilon
\end{cases}$ \\
\tcc{Note the action selection here is not based on the mean of the value distribution}
Execute $a_t$, get reward $r_t$ and next state $s_{t+1}$\\
$a^* \leftarrow \arg\max_{a^\prime}\sum_{i=1}^N q_i(s_{t+1}, a^\prime)$ \\
$y_{t, i} \leftarrow r_t + \gamma q_i(s_{t+1}, a^*)$ for $i=1, \dots, N$\\
$L_\theta \leftarrow \frac{1}{N}\sum_{i^\prime=1}^N \sum_{i=1}^N \Big[\rho_{\hat{\tau}_i}^\kappa\big(y_{t, i^\prime} - q_i(s_t, a_t)\big)\Big]$ \\
$y \leftarrow \beta\max_{\omega^\prime}Q_\Omega(s_{t+1}, \omega^\prime) + (1 - \beta)Q_\Omega(s_{t+1}, \omega_t)$ \\
$L_\psi \leftarrow \frac{1}{2}(r_t + \gamma y - Q_\Omega(s_t, \omega_t))^2$ \\
$\theta \leftarrow \theta - \alpha \nabla_\theta L_\theta$\\
$\psi \leftarrow \psi - \alpha \nabla_\psi L_\psi$\\
}
\caption{\label{alg:qo-d}QUOTA for discrete-action problems}
\end{algorithm}
The algorithm is presented in an online form for simplicity. Implementation details are illustrated in the next section.

\subsection{Experimental Results in Atari Games}
DQN used experience replay to stabilize the training of the convolutional neural network function approximator. However, a replay buffer storing pixels consumes much memory, and the training of DQN is slow. \citet{mnih2016asynchronous} proposed asynchronous methods to speed up training, where experience replay was replaced by multiple asynchronous workers. Each worker has its own environment instance and its own copy of the learning network. Those workers interact with the environments and compute the gradients of the learning network in parallel in an asynchronous manner. Only the gradients are collected by a master worker. This master worker updates the learning network with the collected gradients and broadcast the updated network to each worker. However, asynchronous methods cannot take advantage of a modern GPU (\citealt{mnih2016asynchronous}). To address this issue, \citet{coulom2006efficient} proposed batched training with multiple synchronous workers. Besides multiple workers, \citet{mnih2016asynchronous} also used $n$-step Q-learning. Recently, the $n$-step extension of Q-learning is shown to be a crucial component of the Rainbow architecture (\citealt{hessel2017rainbow}), which maintains the state-of-the-art performance in Atari games. The $n$-step Q-learning with multiple workers has been widely used as a baseline algorithm (\citealt{oh2017value}; \citealt{farquhar2018treeqn}). In our experiments, we implemented both QUOTA for discrete-action control and QR-DQN with multiple workers and an $n$-step return extension.

\begin{figure}
\centering
\includegraphics[width=0.9\textwidth]{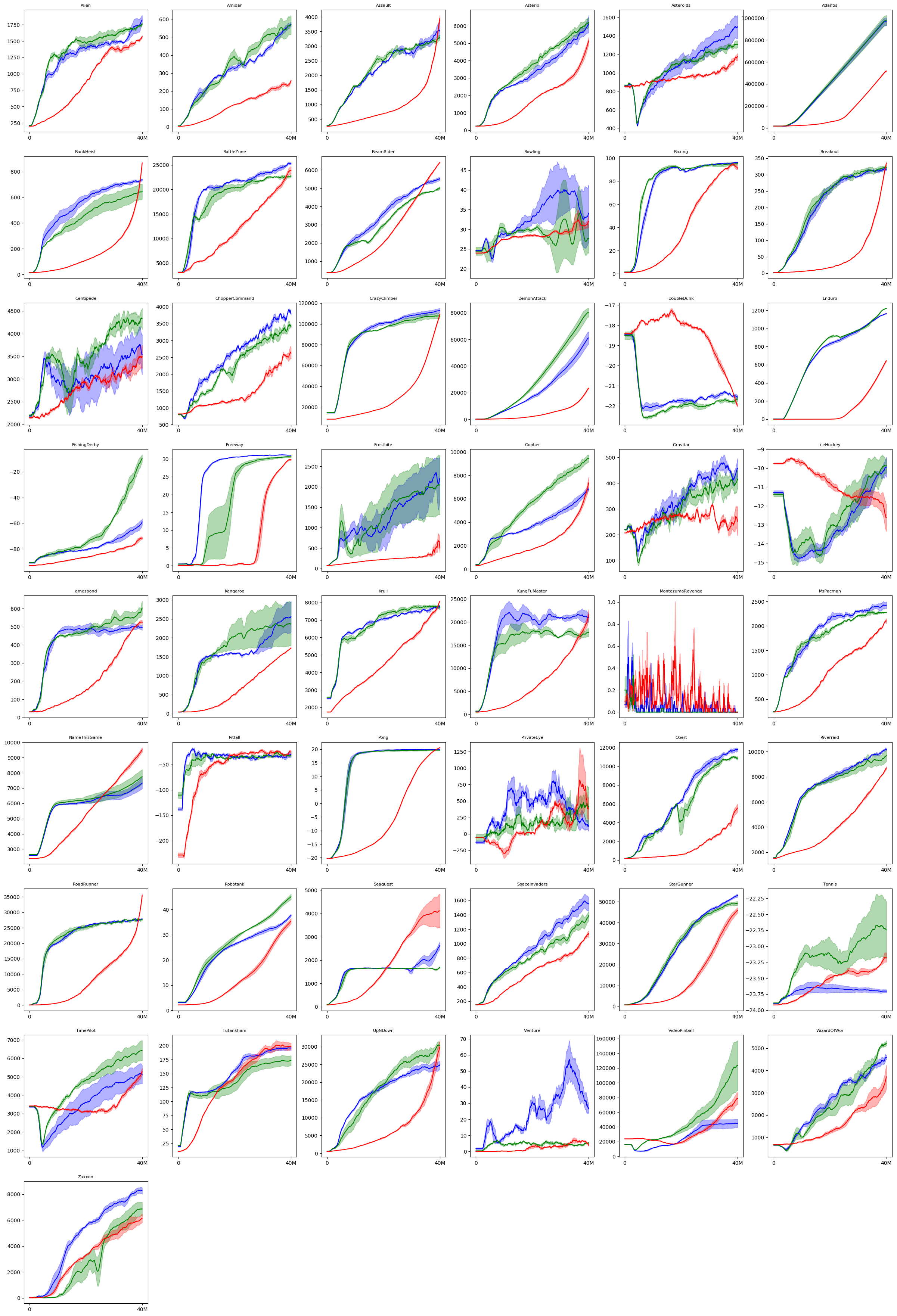}
\caption{\label{fig:atari-all} Learning curves of the 49 Atari games. The x-axis is training steps, and the y-axis is the mean return of the most recent 1,000 episodes up to the time step. Blue lines indicate {\color{blue} QUOTA}, green lines indicate {\color{green} QR-DQN}, and red lines indicate {\color{red} QR-DQN-Alt}. All the scores are averaged over 3 independent runs, and standard errors are plotted as shadow.}
\end{figure}

\begin{table}
\begin{center}
\begin{tabular}{ |c|c|c|c| } 
\hline
\textbf{Game} & \textbf{QUOTA} & \textbf{QR-DQN} & \textbf{QR-DQN-Alt}\\
\hline
Alien & \textbf{1821.91} & 1760.00 & 1566.83 \\
Amidar & \textbf{571.46} & 567.97 & 257.50 \\
Assault & 3511.17 & 3308.78 & \textbf{3948.54} \\
Asterix & 6112.12 & \textbf{6176.05} & 5135.85 \\
Asteroids & \textbf{1497.62} & 1305.30 & 1173.98 \\
Atlantis & 965193.00 & \textbf{978385.30} & 516660.60 \\
BankHeist & 735.27 & 644.72 & \textbf{866.66} \\
BattleZone & \textbf{25321.67} & 22725.00 & 23858.00 \\
BeamRider & 5522.60 & 5007.83 & \textbf{6411.33} \\
Bowling & \textbf{34.08} & 27.64 & 31.98 \\
Boxing & \textbf{96.16} & 95.02 & 91.12 \\
Breakout & 316.74 & 322.17 & \textbf{334.45} \\
Centipede & 3537.92 & \textbf{4330.31} & 3492.00 \\
ChopperCommand & \textbf{3793.03} & 3421.10 & 2623.57 \\
CrazyClimber & \textbf{113051.70} & 107371.67 & 109117.37 \\
DemonAttack & 61005.12 & \textbf{80026.68} & 23286.76 \\
DoubleDunk & \textbf{-21.56} & -21.66 & -21.97 \\
Enduro & 1162.35 & \textbf{1220.06} & 641.76 \\
FishingDerby & -59.09 & \textbf{-9.60} & -71.70 \\
Freeway & \textbf{31.04} & 30.60 & 29.78 \\
Frostbite & \textbf{2208.58} & 2046.36 & 503.82 \\
Gopher & 6824.34 & \textbf{9443.89} & 7352.45 \\
Gravitar & \textbf{457.65} & 414.33 & 251.93 \\
IceHockey & -9.94 & \textbf{-9.87} & -12.60 \\
Jamesbond & 495.58 & \textbf{601.75} & 523.37 \\
Kangaroo & \textbf{2555.80} & 2364.60 & 1730.33 \\
Krull & 7747.51 & 7725.47 & \textbf{8071.42} \\
KungFuMaster & 20992.57 & 17807.43 & \textbf{21586.23} \\
MontezumaRevenge & \textbf{0.00} & 0.00 & 0.00 \\
MsPacman & \textbf{2423.57} & 2273.35 & 2100.19 \\
NameThisGame & 7327.55 & 7748.26 & \textbf{9509.96} \\
Pitfall & -30.76 & -32.99 & \textbf{-25.22} \\
Pong & 20.03 & 19.66 & \textbf{20.59} \\
PrivateEye & 114.16 & \textbf{419.35} & 375.61 \\
Qbert & \textbf{11790.29} & 10875.34 & 5544.53 \\
Riverraid & \textbf{10169.87} & 9710.47 & 8700.98 \\
RoadRunner & 27872.27 & 27640.73 & \textbf{35419.80} \\
Robotank & 37.68 & \textbf{45.11} & 35.44 \\
Seaquest & 2628.60 & 1690.57 & \textbf{4108.77} \\
SpaceInvaders & \textbf{1553.88} & 1387.61 & 1137.42 \\
StarGunner & \textbf{52920.00} & 49286.60 & 45910.30 \\
Tennis & -23.70 & \textbf{-22.74} & -23.17 \\
TimePilot & 5125.13 & \textbf{6417.70} & 5275.07 \\
Tutankham & 195.44 & 173.26 & \textbf{198.89} \\
UpNDown & 24912.70 & \textbf{30443.61} & 29886.25 \\
Venture & \textbf{26.53} & 5.30 & 3.73 \\
VideoPinball & 44919.13 & \textbf{123425.46} & 78542.20 \\
WizardOfWor & 4582.07 & \textbf{5219.00} & 3716.80 \\
Zaxxon & \textbf{8252.83} & 6855.17 & 6144.97 \\
\hline
\end{tabular}
\end{center}
\caption{\label{tab:atari} Final scores of 49 Atari games. Scores are the averaged episode return of the last 1,000 episodes in the end of training. All the scores are averaged over 3 independent runs.}
\end{table}

\newpage
\subsection{Quantile Regression DDPG}
\begin{algorithm}
\DontPrintSemicolon
\textbf{Input:}\\
$\epsilon$: a noise process \\
$\{q_i\}_{i=1,\dots,N}$: quantile estimation functions, parameterized by $\theta$ \\
$\mu$: a deterministic policy, parameterized by $\phi$ \\
\textbf{Output:}\\
parameters $\theta$, $\phi$\\
\hrulefill\\
\For{each time step t} {
Observe the state $s_t$ \\
$a_t \leftarrow \mu(s_t) + \epsilon_t$ \\
Execute $a_t$, get reward $r_t$ and next state $s_{t+1}$\\
$a^* \leftarrow \mu(s_{t+1})$ \\
$y_i \leftarrow r_t + \gamma q_i(s_{t+1}, a^*)$, for $i=1, \dots, N$\\
$L_\theta \leftarrow \frac{1}{N} \sum_{i=1}^N \sum_{i^\prime=1}^N \Big[\rho_{\hat{\tau}_i}^\kappa\big(y_{i^\prime} - q_i(s_t, a_t)\big)\Big]$ \\
$\theta \leftarrow \theta - \alpha \nabla_\theta L_\theta$\\
$\phi \leftarrow \phi + \alpha \nabla_a \frac{1}{N}\sum_{i=1}^N q_i(s_t, a)|_{a=\mu(s_t)} \nabla_\phi \mu(s_t)$ \\
}
\caption{\label{alg:qr-ddpg}QR-DDPG}
\end{algorithm}

The algorithm is presented in an online learning form for simplicity. But in our experiments, we used experience replay and a target network same as DDPG.

\newpage
\subsection{QUOTA for Continuous-Action Problems}
\begin{algorithm}
\DontPrintSemicolon
\textbf{Input:}\\
$\epsilon$: a noise process \\
$\epsilon_\Omega$: random option selection probability \\
$\beta$: option termination probability \\
$\{q_i\}_{i=1,\dots,N}$: quantile estimation functions, parameterized by $\theta$ \\
$\{\mu^i\}_{i=0, \dots, M}$: quantile actors, parameterized by $\phi$ \\
$Q_\Omega$: option value function, parameterized by $\psi$  \\ 
\textbf{Output:}\\
parameters $\theta$, $\phi$, and $\psi$ \\
\hrulefill\\
\For{each time step t} {
Observe the state $s_t$ \\
Select a candidate option $\omega_t$ from $\{\omega^0, \dots, \omega^M\}$\\
$\omega_t \leftarrow \begin{cases}
\omega_{t-1} & \text{w.p. } 1 - \beta \\
\text{random option} & \text{w.p. } \beta \epsilon_\Omega \\
\arg\max_\omega Q_\Omega(s_t, \omega)  & \text{w.p. } \beta (1 - \epsilon_\Omega) 
\end{cases}$ \\
Get the quantile actor $\mu_t$ associated with the option $\omega_t$ \\ 
$a_t \leftarrow \mu_t(s_t) + \epsilon_t$ \\
Execute $a_t$, get reward $r_t$ and next state $s_{t+1}$\\
$a^* \leftarrow \mu_0(s_{t+1})$ \\
\tcc{The quantile actor $\mu_0$ is to maximize the mean return and is used for computing the update target for the critic.}
$y_{t, i} \leftarrow r_t + \gamma q_i(s_{t+1}, a^*)$, for $i=1, \dots, N$\\
$L_\theta \leftarrow \frac{1}{N} \sum_{i=1}^N \sum_{i^\prime=1}^N \Big[\rho_{\hat{\tau}_i}^\kappa\big(y_{t, i^\prime} - q_i(s_t, a_t)\big)\Big]$ \\
$y \leftarrow \beta\max_{\omega^\prime}Q_\Omega(s_{t+1}, \omega^\prime) + (1 - \beta)Q_\Omega(s_{t+1}, \omega_t)$ \\
$L_\psi \leftarrow \frac{1}{2}(r_t + \gamma y - Q_\Omega(s_t, w_t))^2$ \\
$\Delta\phi \leftarrow 0$ \\
\For{$j = 1, \dots, M$} {
$\Delta\phi \leftarrow \Delta\phi + \nabla_a \frac{1}{K}\sum_{k=(j-1)K + 1}^{(j-1)K + K} q_k(s_t, a)|_{a=\mu^j(s_t)} \nabla_\phi\mu^j(s_t)$
}
$\Delta\phi \leftarrow \Delta\phi + \nabla_a \frac{1}{N}\sum_{i=1}^N q_i(s_t, a)_{a=\mu^0(s_t)}\nabla_\phi \mu^0(s_t)$\\
$\phi \leftarrow \phi + \alpha \Delta \phi$ \\
$\theta \leftarrow \theta - \alpha \nabla_\theta L_\theta$\\
$\psi \leftarrow \psi - \alpha \nabla_\psi L_\psi$\\
}
\caption{\label{alg:qo-c}QUOTA for continuous-action problems}
\end{algorithm}
The algorithm is presented in an online learning form for simplicity. But in our experiments, we used experience replay and a target network same as DDPG.

\newpage
\subsection{Experimental Results in Roboschool}
\begin{figure}[h]
\centering
\includegraphics[width=1.0\textwidth]{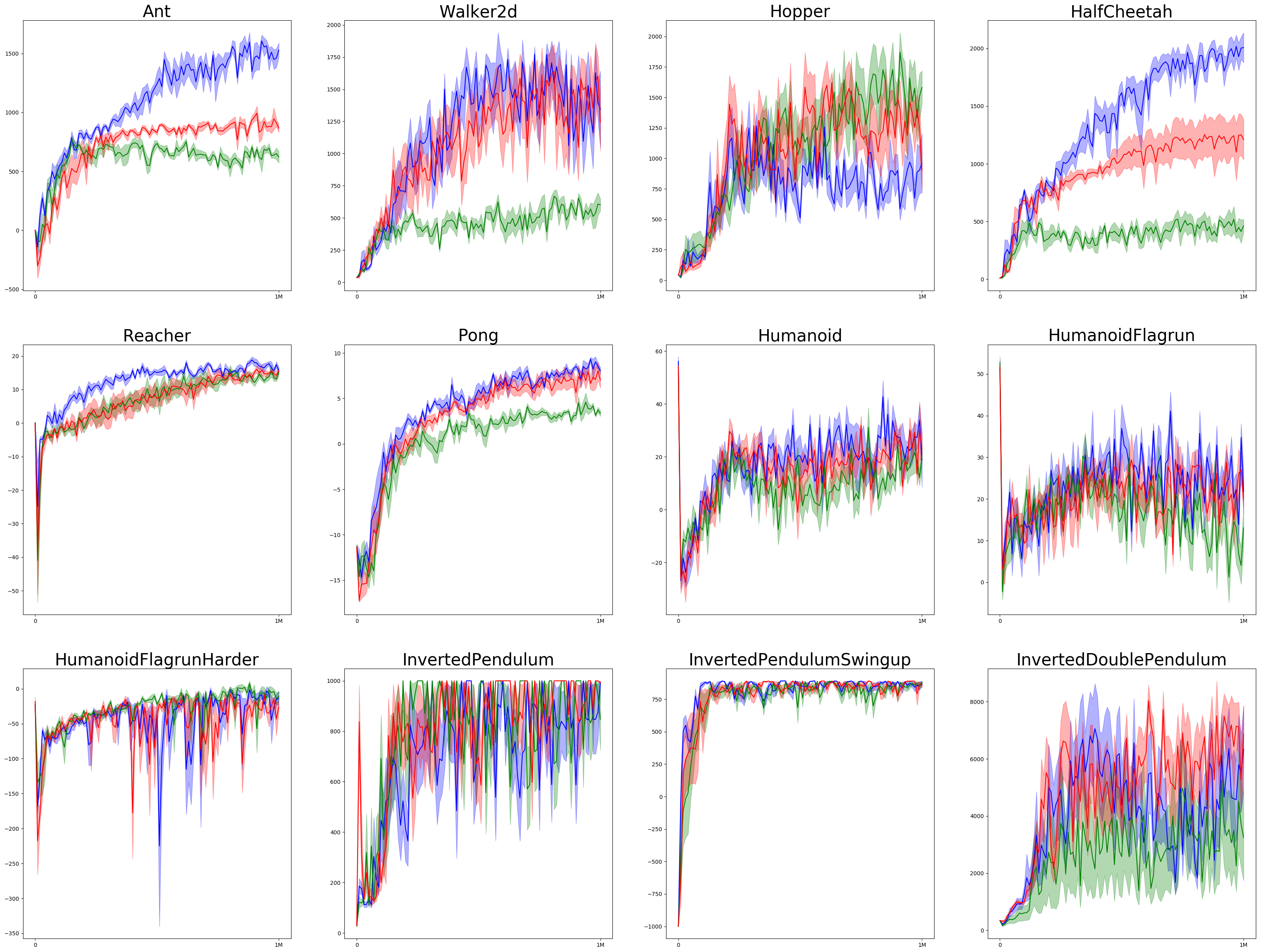}
\caption{Evaluation curves of Roboschool tasks. The x-axis is training steps, and the y-axis is score. Blue lines indicate {\color{blue} QUOTA}, green lines indicate {\color{green} DDPG}, and red lines indicate {\color{red} QR-DDPG}. The curves are averaged over 5 independent runs, and standard errors are plotted as shadow.}
\end{figure}

\end{document}